\documentclass[runningheads]{llncs}
\usepackage[T1]{fontenc}

\usepackage{hyperref}
\usepackage{xcolor}
\usepackage{pdfpages}
\usepackage{graphicx,verbatim}

\usepackage{graphicx}
\graphicspath{ {./images/} }
\usepackage{amsmath}
\usepackage{amssymb}
\begin{document}
\title{NERO: Explainable Out-of-Distribution Detection with Neuron-level Relevance}
\author{
    Anju Chhetri$^{1}$ \and
    Jari Korhonen$^{3}$ \and
    Prashnna Gyawali$^{2}$ \and
    Binod Bhattarai$^{3}$
}
\authorrunning{Chhetri et al.}
\titlerunning{NERO: Explainable OOD Detection with Neuron-level Relevance}
\institute{ $^{1}$NepAl Applied Mathematics and Informatics Institute for research, Nepal\\
    $^{2}$West Virginia University, USA\\
    $^{3}$University of Aberdeen,  Aberdeen, UK\\}

\maketitle              
\begin{abstract}    
Ensuring reliability is paramount in deep learning, particularly within the domain of medical imaging, where diagnostic decisions often hinge on model outputs.
The capacity to separate out-of-distribution (OOD) samples has proven to be a valuable indicator of a model’s reliability in research.
In medical imaging, this is especially critical, as identifying OOD inputs can help flag potential anomalies that might otherwise go undetected.
While many OOD detection methods rely on feature or logit space representations, recent works suggest these approaches may not fully capture OOD diversity. To address this, we propose a novel OOD scoring mechanism, called NERO, that leverages neuron-level relevance at the feature layer. Specifically, we cluster neuron-level relevance for each in-distribution (ID) class to form representative centroids and introduce a relevance distance metric to quantify a new sample’s deviation from these centroids, enhancing OOD separability.
Additionally, we refine performance by incorporating scaled relevance in the bias term and combining feature norms. 
Our framework also enables explainable OOD detection. 
We validate its effectiveness across multiple deep learning architectures on the gastrointestinal imaging benchmarks Kvasir and GastroVision, achieving improvements over state-of-the-art OOD detection methods.


\end{abstract}

\section{Introduction}

        Deep learning-based systems hold significant potential for enhancing computer-aided diagnostics \cite{chan2020deep} in medical image analysis.
        However, most of the deep learning models are trained under the closed-world assumption \cite{he2015delving,krizhevsky2012imagenet}, where the training and testing data share the same distribution.
        This assumption limits their effectiveness in real-world deployment, as these models often encounter out-of-distribution (OOD) data and may produce high-confidence predictions on such data \cite{nguyen2015deep,tang2021codes}.
        For instance, a deep learning model trained on common gastrointestinal conditions in endoscopic images may misclassify rare diseases like blue rubber bleb nevus syndrome (BRBNS) or gastrointestinal stromal tumors (GISTs), which were absent from its training data.
        In such scenarios, the model may erroneously assign high confidence to incorrect predictions, leading to potential misdiagnoses.
        To ensure the trustworthiness of these systems, they must be capable of detecting and flagging OOD data so that the human expert could handle it safely.
        Similarly, the challenge of obtaining labeled datasets, particularly for rare pathologies, results in a long-tailed class distribution.
        This has driven interest in applying OOD detection techniques for pathology identification \cite{tschuchnig2021anomaly,hong2024out}, where healthy data are modeled as in-distribution (ID), while unhealthy data are treated as OOD samples, making OOD detection increasingly relevant in medical deep learning.

        In this work, we focus on post-hoc OOD detection, an unsupervised approach that does not require any model re-training or access to OOD data during training, making it highly practical for real-world deployment.
        These methods can generally be categorized into three types.
        Firstly, \textit{logit-based methods}, uses information from the model’s logits.
        In \cite{hendrycks2016baseline}, probabilities from softmax layers are utilized, while \cite{liu2020energy,hendrycks2019scaling} refine this method by computing energy scores on the logits and using maximum logit value for OOD detection, respectively.
        Secondly, the \textit{feature-based methods}, focuses on analyzing feature representations extracted from the model.
        In \cite{lee2018simple}, Mahalanobis distance between feature vectors and their class-wise mean is computed, while \cite{sun2021react} introduce activation rectification in the penultimate layer to enhance OOD detection.
        Similarly, in \cite{pokhrel2024ncdd}, the OOD score is computed as a weighted difference between the sum of distances to non-nearest centroids and the distance to the nearest centroid.
        Finally, the \textit{gradient-based methods} include \cite{huang2021importance}, where GradNorm uses vector norm of gradients, backpropagated  from the KL divergence between the softmax output and a uniform probability.
        Recent works also advocate for the integration of multiple information to enhance OOD detection. For instance, \cite{wang2022vim} introduces a virtual logit that utilizes information from features, logits, and probabilities for OOD scoring.

        Fundamentally a different perspective, and building on the observation that deep neural networks often assign specialized roles to individual neurons for particular classes \cite{bau2017network,olah2018building}, we present, for the first time, an approach that harnesses neuron-level relevance for OOD detection-\textit{NERO}.
        Specifically, we cluster neuron-level relevance for samples in each class in the ID data to form representative centroids. We then introduce a novel \textit{relevance distance} metric, which quantifies how far a new sample’s relevance signature is from these class-centric centroids, thus serving as an effective OOD separability function.
        Unlike methods focusing purely on activation magnitudes or feature vectors, our approach considers how neurons collectively contribute to the final prediction.
        Additionally, we incorporate the scaled relevance in the bias term and combine feature norms to further refine OOD detection performance.
        Furthermore, since neuron-level relevance has already demonstrated its utility in explainable AI \cite{montavon2019layer,horst2019explaining,binder2018towards}, our framework naturally extends to explainable OOD detection—an especially critical feature in sensitive domains like medical imaging, where both reliability and interpretability are paramount.
        We evaluated our method on two different gastrointestinal datasets and two separate classifier backbones, ResNet-18 \cite{he2016deep} and Data-efficient Image Transformer (DeiT) \cite{touvron2021training}, demonstrating its superior performance compared to existing methods.
        Overall, our contributions are as follows:

        \begin{enumerate}    
            \item We propose a novel explainable post-hoc OOD detection method, NERO, that leverages neuron-level relevance to analyze prediction-relevant patterns.
            \item We demonstrate the effectiveness of our method through extensive empirical evaluations on challenging medical image benchmarks, showing superior performance compared to state-of-the-art OOD detection techniques.
        \end{enumerate}

\section{Method}

    \noindent \textbf{Problem Formulation}: We consider a multi-class classification task, where we represent training and testing sets as \begin{math} \mathcal{D}_t = \{(x_i, y_i)\}_{i=0}^{n_t} \end{math} and  \begin{math} \mathcal{D}_e = \{(x_i, y_i)\}_{i=0}^{n_e}\end{math} respectively.
         $x_i$  represents an image and $y_i \in$ \{0,...,C-1\} is its corresponding class label, where C is the total number of classes.
         The data in $\mathcal{D}_t$ and $\mathcal{D}_e$ are assumed to be independent and identically distributed (iid) and 
         are drawn from $\mathbb{P}_{\text{id}}$.
        We use $g$ to represent a neural network trained on $\mathcal{D}_t$, parameterized by $w$.
        The feature extractor $f_w(\cdot)$ is derived from the penultimate layer of $g$.
        The aim of OOD detection is to derive a confidence score 
        that allows us to assess whether a given test data ($x$) is drawn from $\mathbb{P}_{\text{id}}$ or from another distribution $\mathbb{P}_{\text{ood}}$ with unknown OOD class. 
        
\begin{figure}[!htb]
    \centering
    \includegraphics[trim= 0cm  0cm 0cm 0cm, width=1\textwidth]{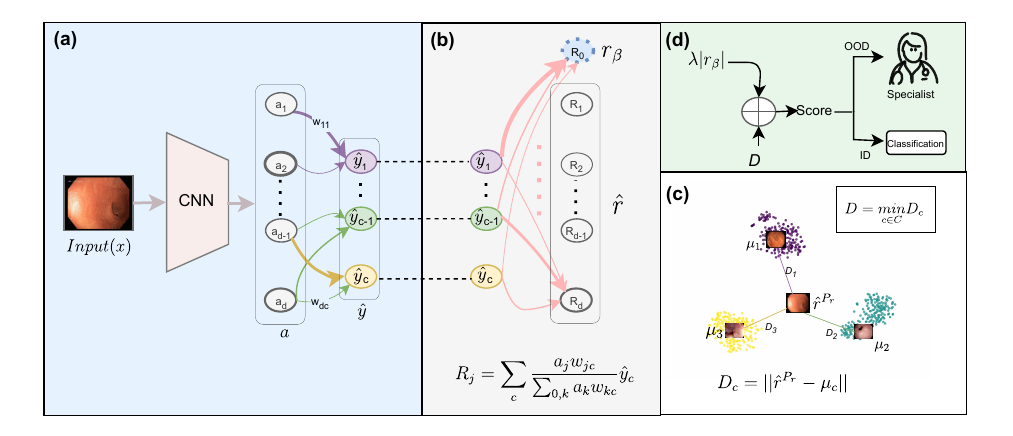} 
    \caption{Illustration of \textbf{NERO}. \textbf{(a)} Feature extraction: Input images are processed through a CNN to obtain feature ($a$) in the penultimate layer and class predictions ($\hat{y}$). \textbf{(b)} Relevance estimation: Neuron contributions are quantified by calculating relevance scores ($R_j$) based on connection strengths, with $R_0$ representing bias contribution. \textbf{(c)} Distance computation: The framework calculates distances ($D_c$) between test sample's relevance and class-wise centroids ($\mu_c$) in principal component space. \textbf{(d)} OOD scoring: The detection mechanism combines the minimum distance between sample relevance patterns and class centroids ($D$) along with the absolute bias relevance term ($|r_\beta|$).}
    \label{fig:method}
\end{figure}

    Our approach focuses on analyzing the contribution of individual neurons to model predictions when processing ID versus OOD samples. For any input $x$, we obtain the network predictions, $\hat{y} = g_w(x)\in \mathbb{R}^C$ and features from the penultimate layer,  $a = f_w(x)\in \mathbb{R}^d$, where $d$ denotes the total number of features (Fig. \ref{fig:method} (a)).
    To quantify the contribution of each neuron to the final prediction, we compute relevance scores for each neuron in the penultimate layer.

    \noindent\textbf{Relevance Estimation}:
    The relevance is computed by summing over all neurons $c$ in the final layer, where each neuron's contribution in the 
    penultimate 
    layer is weighted by the ratio of the specific connection's strength $(a_jw_{jc})$ to the total input received by neuron $c$ $(\sum a_kw_{kc})$ (Fig.\ref{fig:method} (b)).
    Higher relevance values indicate neurons that substantially influence the prediction outcome. 
    \noindent For relevance calculation, we follow the definition of LRP-0 \cite{bach2015pixel,montavon2019layer}:

     \begin{equation}
        \hat{r}_j(x) = \sum_c \frac{a_j w_{jc}}{\sum_{0,k} a_k w_{kc}} \hat{y}_c, \forall j \in \{0,..,d\} 
    \end{equation}

        \noindent An additional neuron $r_\beta(x) \in \mathbb{R} $ is introduced to represent the bias contribution, with $a_0$=1 and and $w_\text{0c}$ denoting the bias value.
        To quantify how individual samples relate to representative class distributions -- a key differentiator for ID examples from OOD examples -- we compute class-wise relevance centroids.

         \noindent\textbf{Mean Relevance Estimation}:
         For centroid calculations, we utilize only the first $d$ features (i.e., neurons) to ensure neuron-wise comparison, while the bias neuron, $r_\beta$, due to their difference in magnitude is handled separately.
         Let $\hat{r}(x) \in \mathbb{R}^d$ represent the neuron-level relevance score of the input $x$ from the $\mathcal{D}_t$, respectively, and we define a matrix $A \in \mathbb{R}^{n_t \times d}$ containing all $\hat{r}(.)$. We observed that these relevance patterns exhibited distinct class-specific characteristics.
         To capture these class-specific patterns, we compute class-wise mean relevance scores from the training set.
         First, to preserve the most informative variations while reducing noise and redundancy, we apply Principal Component Analysis (PCA) to $A$ and decompose it into principal space, defined by projection of a matrix $P_r\in \mathbb{R}^{d \times z}$, spanned by the first z columns.
         
    \begin{equation}
        \mu_c = \frac{1}{n_c} \sum_{i=1}^{n_c}  \hat{r}(x_i)^{P_r}
    \end{equation}
    where, $r(.)^{P_r}$ is the projection of $r(.)$ to $P_r$ and $n_c$ is the total number total number of samples belonging to class c.

    \noindent\textbf{Proposed OOD Score}:
    With the mean relevance obtained, for any new sample $x$, we define the OOD score as: 
    
            \begin{equation} 
                S(x) = \min_{c \in \mathcal{C}} ||\hat{r}(x)^{P_r} - \mu_c|| + \lambda|r_\beta(x)|
            \end{equation}
            Our scoring function incorporates two complementary components: the minimum distance between projected relevance patterns and class-wise means, and the weighted bias relevance term. 
            The minimum distance measures the deviation of relevance patterns from learned class patterns, with smaller distances for ID samples and larger ones for OOD samples (Fig.\ref{fig:method} (c)).
            Through empirical observations, we found that relevance scores distributed in the bias nodes exhibit strong discriminative power for OOD detection.
            Capitalizing on this, we introduce the bias relevance term, weighted by $\lambda$, computed as the ratio of their mean values over the training set, to enhance detection capability while ensuring both terms contribute equally and remain comparable in scale (Fig.\ref{fig:method} (d)). 
            
        \begin{equation}
            \lambda = \frac{\mathbb{E}_{j\in \mathcal{D}_t}[||\hat{r}(x_j)^{P_r}- \mu_c||]}{\mathbb{E}_{j\in \mathcal{D}_t}[|r_{\beta}(x_j)|]}
        \end{equation}
        
         \noindent Building on the insights of previous works like \cite{cook2020outlier}, which utilizes information from the null space of the weight matrix for outlier detection, we analyze neurons with lower relevance scores to capture subtle OOD attributes.
         A model trained on ID data primarily captures ID-specific features, with highly relevant neurons focusing on these characteristics. Introducing a scaling term that considers the norm of normalized features ($\hat{f}_w(.)$) from the less relevant neurons ensures that informative OOD signals are not overshadowed by dominant ID features. This refinement enhances the model’s ability to capture OOD attributes, improving performance in distinguishing OOD from ID data.
         
         \begin{equation}
         \label{eq:finalScore}
         S(x) = (\min_{c \in \mathcal{C}} ||\hat{r}(x)^{P_r} - \mu_c|| + \lambda|r_\beta(x)|)\cdot(\sum_{j\in B_k}{|\hat{f}_w(x)_j|})
        \end{equation}
        where $B_k$ = $\{ j \mid j \in \operatorname{argmin}_{J \subseteq [d], |J| = k} |\hat{r}_j(x)| \}$ is the set of indices corresponding to the $k$ least relevant neurons.
        With $S(\cdot)$ defined in Eqn \ref{eq:finalScore}, OOD detection is framed as a binary classification task with a threshold $\lambda_\text{S}$, that guides the decision process with those exceeding the threshold as OOD and the rest as ID. The threshold $\lambda_\text{S}$ is often set for a 95\% true positive rate on test set.

\section{Experiments and Results}

    \textbf{Datasets and Setup:} We evaluated our method using two publicly available multi-class endoscopy datasets: Kvasir-v2 \cite{sharma2023deep} and GastroVision \cite{jha2023Gastrovision}. These datasets provide comprehensive benchmarking in real-world settings with both balanced and unbalanced class distributions.
The Kvasir-v2 dataset comprises 8,000 gastrointestinal tract images equally distributed across 8 classes (1,000 per class), representing pathological findings, anatomical landmarks, and endoscopic procedures.
We designated three healthy anatomical landmarks-Pylorus, Z-line, and Cecum-as ID classes, with the remaining five serving as OOD classes corresponding to abnormalities.
The GastroVision dataset encompasses 27 classes across upper and lower GI categories, totaling 8,000 images. The class distribution is imbalanced, with samples per class ranging from 6 to 1,467.
We categorized 11 normal and anatomical findings as ID data, while pathological and therapeutic findings were designated as OOD samples. For both datasets, we used an 80:20 train-test split, resulting in 800 training and 200 test images per class for Kvasir-v2 (2,400 training, 600 test total). For GastroVision, this yielded 3,804 training and 955 test images across all classes.

    \noindent\textbf{Implementation Details}:
    We experiment with both CNN and transformer architectures, using ResNet-18 and DeiT \cite{touvron2021training} for image classification.
For ResNet-18, we extract 512-dimensional feature vectors from the global average pooling layer before the final classification layer. For the transformer, we use the small DeiT model (DeiT-S) with a 16×16 patch size, where the [CLS] token embedding from the final transformer block (384-dimensional) serves as the feature representation.
We initialize both models with ImageNet pre-trained weights and fine-tune them for classification. 
The models are optimized using Adam \cite{kingma2014adam} with a 1$e^\text{-4}$ learning rate.
Input images from Kvasir-v2 and GastroVision are resized to 224×224 pixels. 
Training was performed on NVIDIA A100 GPUs for 20 epochs on Kvasir-v2 and 50 on GastroVision.
    
    \noindent\textbf{Evaluation Metrics:} 
    We evaluate our model using AUROC and FPR95.
    AUROC measures the overall performance of the model in distinguishing between classes. It is threshold-independent, with higher values indicating better performance.
    FPR95 quantifies the false positive rate when the true positive rate reaches 95\%, with lower values indicating superior performance.
    
    \noindent\textbf{Baselines:} We  compared our method with 10 competitive methods MSP \cite{hendrycks2016baseline}, ODIN \cite{liang2017enhancing}, Energy \cite{liu2020energy}, Entropy \cite{chan2021entropy}, MaxLogit \cite{hendrycks2019scaling}, NECO \cite{ammar2023neco}, ViM \cite{wang2022vim}, ReAct\cite{sun2021react}, GradNorm \cite{huang2021importance}, and Mahalanobis following this setting\cite{lee2018simple}.

\subsection{Quantitative Results}
\noindent\textbf{NERO vs. Baselines}: Table \ref{results} presents 
our main results on ResNet-18 and DeiT for the Kvasir and GastroVision datasets.
The best results are highlighted in bold, while the second-best results are underlined.
NERO demonstrates competitive, or even superior, performance compared to state-of-the-art approaches.
On Kvasir-v2, it achieves the lowest FPR95 (28.84\% with ResNet-18, 18.96\% with DeiT), outperforming all baselines.
It also ranks among the top three in AUROC (90.76\% with ResNet-18, 92.73\% with DeiT).
On GastroVision, NERO delivers the best results on both metrics with DeiT and second best with ResNet-18.
Its consistent performance across different architectures and datasets highlights its robustness in reducing false positives while maintaining high detection accuracy.

\begin{table}[!t]
\caption{Performance comparison of different methods using ResNet-18 and DeiT on Kvasir-v2 and GastroVision. Metrics include AUROC (higher is better, $\uparrow$) and FPR95\% (lower is better, $\downarrow$), both reported as percentages.}
\centering
\begin{tabular}{l c c c c c c c c}
\hline
&\multicolumn{4}{c}{ResNet-18} & \multicolumn{4}{c}{DeiT} \\
&\multicolumn{2}{c}{Kvasir-v2} & \multicolumn{2}{c}{GastroVision} & \multicolumn{2}{c}{Kvasir-v2} & \multicolumn{2}{c}{GastroVision} \\
Method & AUC & FPR95 & AUC & FPR95 & AUC & FPR95 & AUC & FPR95 \\
\hline
MSP &90.3 &41.72 &66.93 &90.56 &87.05 &40.18 &70.0 &90.74  \\
ODIN  &\textbf{91.77} &\underline{35.44}  &69.79 &79.27 &88.41 &36.4 &73.37 &83.68  \\
Energy  &88.85 &52.36  &70.31 &79.79 &85.77 &44.02 &75.35 & 83.68 \\
Entropy &90.38 &41.86  &67.37 &87.32 &87.2 &39.94 &70.34 &90.19 \\
MaxLogit &88.9 &52.38  &70.08 &80.44 &85.77 &44.02 &75.11 &84.2 \\
Mahalanobis &84.05  &54.06  &65.93  &89.69 &\textbf{94.50}  &\underline{21.86} &75.68 &81.43  \\
ViM &90.62 &41.1  &72.70 &76.98 &\underline{93.88}  &24.38 & 76.69 &\underline{78.37} \\
NECO &89.64 &47.90  &\textbf{79.81}  &\textbf{71.61} &88.31  &37.60 & \underline{76.95} &81.92 \\
Energy+ReAct &86.57 &53.78  &61.93  &83.86 &83.49  &46.84 &73.42  &83.22 \\
GradNorm &85.33 &54.68 &62.55 & 90.5 &71.33  &57.8 &54.85 &88.68 \\
\textbf{NERO (ours)} &\underline{90.76} &\textbf{28.84} &\underline{75.95} &\underline{74.33}  &92.73  &\textbf{18.96} &\textbf{82.03} &\textbf{76.74} \\

\hline
\end{tabular}
\label{results}
\end{table}

\begin{figure}[!t]
    \centering
    \includegraphics[width=1\textwidth]{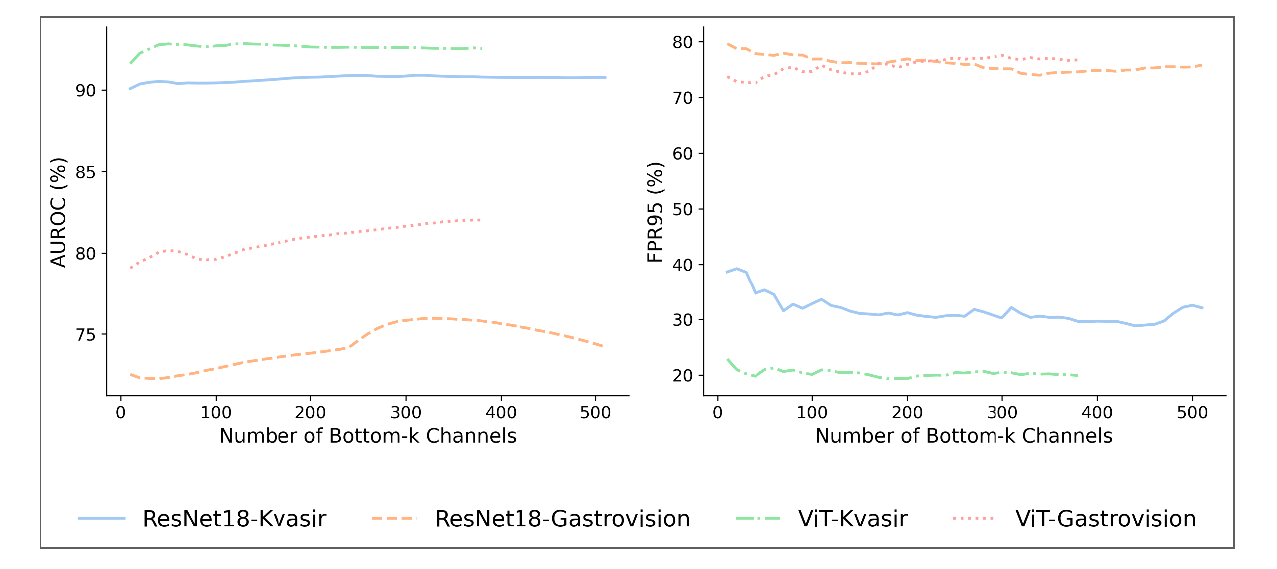} 
    \caption{Robustness to the number of bottom channels selected, with AUROC shown on the left and FPR95 on the right. }
    \label{fig:Robustness to k}
    \end{figure}

    \noindent\textbf{The effect of hyperparameter:}
    Our results show that performance remains robust across different numbers of bottom channels selected for feature scaling. As seen in Fig. \ref{fig:Robustness to k}, metrics stay stable over a broad range, though using very few low-relevance channels can be suboptimal, as indicated by the relevance distribution.

\subsection{Qualitative Results}
In this section, we demonstrate the explainability of our proposed OOD framework. To this end, we consider two cases comparing ID and OOD samples. For each category, we visualize the specific features that neurons attend to when processing ID and OOD samples.
To provide visual insights into how the OOD detector distinguishes between ID and OOD data, we select the top four channels with the highest relevance scores for each ID class and visualize their relevance maps on both correctly classified ID samples and OOD samples. For visualization, we use Concept Relevance Propagation (CRP) \cite{achtibat2023attribution} to compute channel-conditional relevance maps $R$(x|y), where x represents the input image and y represents a set of conditions.

\begin{figure}[!t]
    \centering
\includegraphics[width=1\textwidth]
{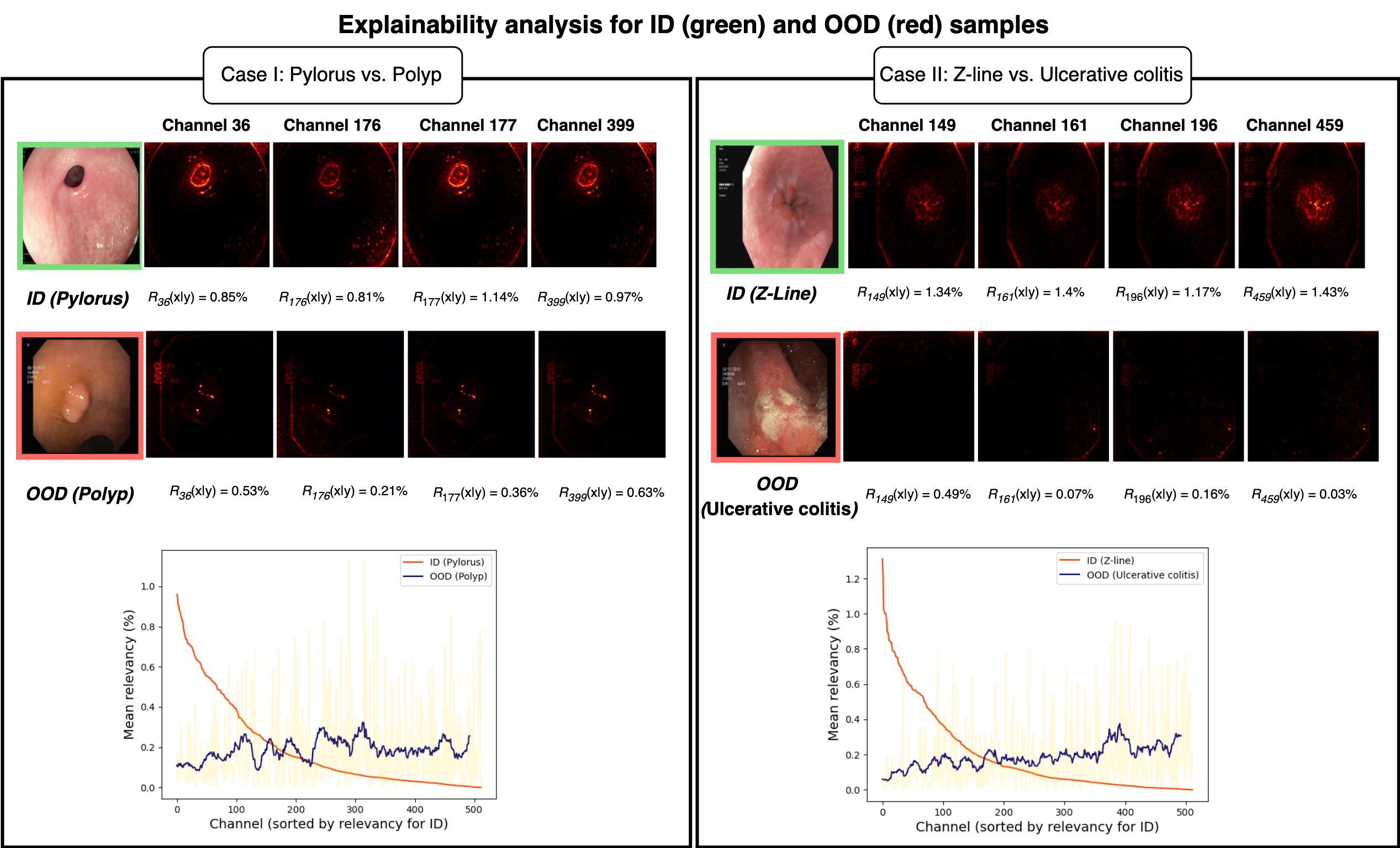} 
    \caption{
Explainability analysis of ID and OOD samples for two cases. In each case, the top two rows show attribution maps for ID and OOD samples across the top four channels with the highest relevance scores (per ID class). The bottom plot depicts the distribution of mean relevance scores across all channels for both ID and OOD samples.
}
    \label{fig:attribution}
\end{figure}

The results are presented in Fig. \ref{fig:attribution}. The left panel illustrates Case I, where we analyze Pylorus (ID class) and Polyp (OOD data), while the right panel depicts Z-line (ID class) and Ulcerative Colitis (OOD data). The top two rows show the corresponding relevance maps for ID and OOD samples, respectively. Notably, our framework successfully flags the second-row samples as OOD, and the visualizations reveal distinct patterns: ID samples exhibit focused, high-intensity activations around key anatomical landmarks, whereas OOD samples
display markedly different attribution patterns, with lower and more diffuse intensity values across the channels.
Further, to demonstrate the consistency of this pattern across our dataset, we present the mean relevance scores for both ID and OOD class plots for each case (Fig. \ref{fig:attribution}, bottom), with channel indices sorted according to the mean relevance score of the ID class. We applied a moving average to the OOD data to highlight the general trend. These plots illustrate a clear distinction between ID and OOD relevance scores across channels.

\section{Conclusion}

In this paper, we introduced Neural Relevance-based OOD detection (NERO), a novel post-hoc OOD detection method that leverages neuron-level relevance patterns to distinguish OOD samples. By clustering neuron-level relevance signatures for each ID class and quantifying the relevance distance of test samples from these class-centric centroids, our approach provides a robust framework for OOD detection. Our evaluations on challenging gastrointestinal datasets demonstrated its effectiveness, achieving consistently superior performance across diverse model architectures, including CNN-based (ResNet-18) and transformer-based (DeiT) networks.

\clearpage
\bibliographystyle{splncs04}
\bibliography{mybibliography}
\end{document}